\def\year{2015}
\documentclass[letterpaper]{article}
\usepackage{aaai}
\usepackage{times}
\usepackage{helvet}
\usepackage{courier}
\usepackage{graphicx} 
\usepackage{float}
\usepackage{color}
\makeatletter
\def\@biblabel#1{}
\makeatother

\usepackage{subfigure}

\begin{document}
%
\title{Privacy Prediction of Images Shared on Social Media Sites Using Deep Features}
\author{Ashwini Tonge and Cornelia Caragea\\
Computer Science and Engineering\\
University of North Texas\\
{\tt ashwinitonge@my.unt.edu, ccaragea@unt.edu}\\
}
\maketitle
\begin{abstract}
\begin{quote}
Online image sharing in social media sites such as Facebook, Flickr, and Instagram can lead to unwanted disclosure and privacy violations, when privacy settings are used inappropriately.
With the exponential increase in the number of images that are shared online every day, the development of {\em effective} and {\em efficient} prediction methods for image privacy settings are highly needed. The performance of models critically depends on the choice of the feature representation. In this paper, we present an approach to image privacy prediction that uses deep features and deep image tags as feature representations. Specifically, we explore deep features at various neural network layers and use the top layer (probability) as an auto-annotation mechanism. 
The results of our experiments show that models trained on the proposed deep features and deep image tags substantially outperform baselines such as those based on SIFT and GIST as well as those that use ``bag of tags'' as features. 
\end{quote}
\end{abstract}

\section{Introduction}
The rapid increase in multi-media sharing through social networking sites such as Facebook, Flickr, and Instagram can cause potential threats to users' privacy. Many users are quick to share private images about themselves, their family and friends, without thinking much about the consequences of an unwanted disclosure of these images. Moreover, social networking sites such as Facebook, allow users to tag other people, which can reveal private information of the users in a particular image \cite{Ahern:2007:OPP:1240624.1240683}. Gross and Acquisti \shortcite{Gross:2005:IRP:1102199.1102214} analyzed more than 4,000 Carnegie Mellon University students' Facebook profiles and outlined potential threats to privacy. Users often provide personal information generously on online social networking websites, but hardly make use of limiting privacy preferences. Additionally, they rarely change default privacy settings, which could jeopardize their privacy \cite{Zerr:2012}. 

Current social networking sites do not assist users in making privacy decisions for images that they upload online. Manually assigning privacy settings to each image each time can be cumbersome. To avoid a possible loss of a user's privacy, it has become critical to develop automated approaches that can accurately predict the privacy settings for images that are shared online. 

Several studies have started to explore classification models of image privacy using image tags and image content features such as SIFT (or Scale Invariant Feature Transform) or RGB (or Red Green Blue) \cite{Zerr:2012,Squicciarini2014} and found that image tags are very informative for the task of classifying images as {\em public} or {\em private}. 
However, given large collections of image training data available these days (e.g., the ILSVRC-2012 subset of the ImageNet dataset \cite{ILSVRC15} that has 1.2M+ images labeled with 1000 categories), recent deep neural networks are now able to learn powerful features that go beyond SIFT and RGB \cite{donahue13,donahue14} and work remarkably well in many image analysis tasks such as generating short sentence descriptions of images and videos \cite{venugopalanXDRM15}.

In this paper, we explore an approach to privacy prediction that uses deep visual features and deep tags for predicting the class of an image as {\em public} or {\em private}. Specifically, we use three deep feature representations corresponding to the output of three layers of an eight-layer deep neural network pre-trained on the above ILSVRC-2012 
\cite{ILSVRC15}, as well as the probability distribution over categories obtained from the last layer of the network via softmax. We further investigate deep tags, which correspond to the top ranked probabilities from the probability distribution over categories. We analyze image tags with respect to privacy settings and use information gain and tag frequency to identify informative tags. 

\vspace{2mm}
{\bf Our Contributions.}
\begin{itemize}
\item We show that models trained using traditional visual features such as ``SIFT'' and ``GIST'' yield very low performance with respect to the private class.
\vspace{-1mm}
\item  We explore deep visual features and deep tags for privacy prediction and show empirically that models trained on these features outperform those trained using SIFT, GIST, and user provided tags.
\vspace{-1mm}
\item We evaluate our approach on Flickr images sampled from the PiCalert dataset (\citeauthor{Zerr:2012} \citeyear{Zerr:2012}). 
\vspace{-1mm}
\item Our tag related analysis can assist in understanding the characteristics of the private and public classes.
\end{itemize}


\section{Related work}
Emerging privacy violations and security threats in social networks have started to attract various researchers to this field. Several works are carried out to study users' privacy concerns in social networks, privacy decisions about sharing resources and the risk associated with them. 

Ahern et al. \shortcite{Ahern:2007:OPP:1240624.1240683} examined privacy decisions and considerations in mobile and online photo sharing. They explored critical aspects of privacy such as users' consideration for privacy decisions, content and context based patterns of privacy decisions, how different users adjust their privacy decisions, and user behavior towards personal information disclosure. They also discussed about the taxonomy of privacy considerations classified into four main themes: security, social disclosure, identity, and convenience. They conclude that applications to support and influence user's privacy decision-making process should be developed. Also, Gross and Aquisti \shortcite{Gross:2005:IRP:1102199.1102214} identified privacy implications and risk associated with it in social networks. 

Buschek et al. \shortcite{Buschek:2015} presented an approach to assigning privacy settings to shared images using metadata (location, time, shot details) and visual features (faces, colors, edges). Zerr et al. \shortcite{Zerr:2012} proposed privacy-aware image classification, as well, in which they learned classifiers trained on Flickr photos. They considered metadata and visual features such as color histograms, faces, edge-direction coherence, SIFT features and average brightness and sharpness for privacy classification task. Squicciarini et al. \shortcite{Squicciarini2014} also applied SIFT for predicting sensitivity of user's images. SIFT \cite{Lowe:2004:DIF:993451.996342} detects scale, rotation and translation invariant key-points of objects in images, and it is widely used for image analysis in computer vision. SIFT extracts a pool of visual feature vectors, which are used to represent a ``bag-of-visual-words.'' We use SIFT as one of our baselines. 

Olivia and Torrance \shortcite{Oliva:2001:MSS:598425.598462} introduced global descriptor features (GIST) for image analysis, and proposed a set of perceptual dimensions (naturalness, openness, roughness, expansion and ruggedness) that represent the dominant spatial structure of the scene. They show that these dimensions can be estimated using spectral and localized information.

Several works were conducted in the context of tag-based access control policies for images \cite{yeung2009,tags12,collcom}, showing some initial success in tying tags with access control rules. However, the  scarcity of tags for many online images \cite{mum12}, and the workload associated with user-defined tags precluded accurate analysis of images'  sensitivity based on this dimension.


As noted, the above approaches majorly involved visual feature such as SIFT, edges-direction coherence, and face detection, for privacy classification. 
Recently, the computer vision community has shifted towards convolutional neural networks for task such as object detection \cite{sermanet:iclr:14,Sermanet:2013:PDU:2514950.2516194} semantic segmentation \cite{farabet-pami-13}. Deep convolutional neural network has also acquired state of the art results on ImageNet (Highly challenging dataset used for object recognition) only using supervised learning \cite{NIPS2012_4824}. Karayev et al. \shortcite{DBLP:journals/corr/KarayevHWAD13} described an approach to predicting the style of images by evaluating different visual features. They found that features learned from multi-layer network generally performed best when trained on 80K Flickr photographs. 

Unlike the previous works described above, we explore the use of deep features and deep tags derived from a  deep convolutional neural network, for privacy prediction of online images. To our knowledge, this is the first work that use deep networks for privacy prediction.  

\begin{figure*}[htb]
\centering
  \begin{tabular}{@{}ccc@{}}
\includegraphics[scale=0.8]{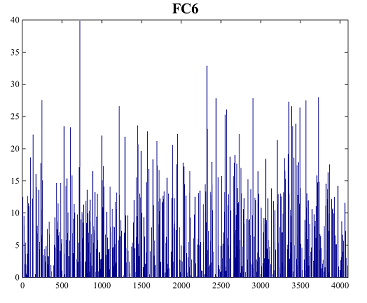} &
\includegraphics[scale=0.6]{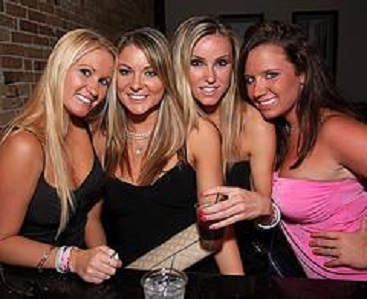} & 
\includegraphics[scale=0.8]{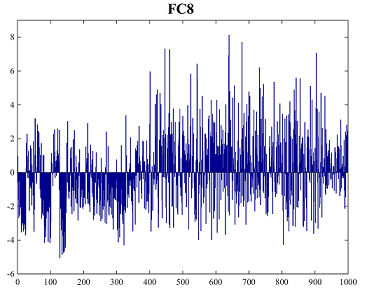} \\
\includegraphics[scale=0.8]{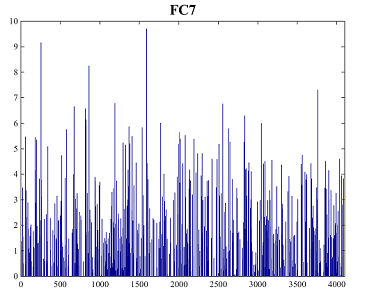} &
\includegraphics[scale=0.6]{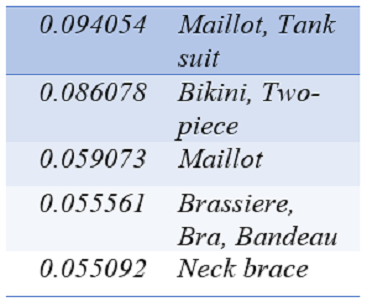} & 
\includegraphics[scale=0.8]{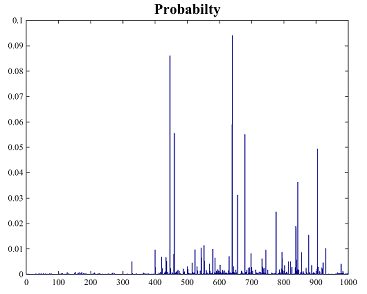} \\  \end{tabular}
\caption{Deep feature representations of a given image.} 
\label{fig:features}
\end{figure*}

\section{Privacy Setting Prediction Using Deep Features and Image Tags}
As discussed earlier, privacy of an image can be determined by the content of that image and the description associated with it. We extracted visual features and tags 
for differentiating between private and public settings. 

\subsection{Image Features:}

\subsubsection{Deep Neural Net (DNN): Deep Features}
The network takes one or more blob as input and produces one or more blob as output. Layers are responsible for forward pass and backward pass. Forward pass takes inputs and generates the outputs. Backward pass takes gradients with respect to the output and computes the gradient with respect to the parameters and to the inputs, which are consecutively back-propagated to the previous layers \cite{Jia:2014:CCA:2647868.2654889}.  In the convolutional neural network architecture, features are extracted from images through each layer in a feed-forward fashion. 
This architecture contains eight layers with weights; the first five are convolutional and the remaining three are fully-connected. 
The three fully connected layers are referred as ``FC$_6$,'' ``FC$_7$,'' and ``FC$_8$,'' and the final output layer is referred as ``Prob.'' ``Prob'' produces a probability distribution over 1000 object category for the input image.
The output of the last fully connected layer is given as input to a 1000-way softmax which produces a distribution over the 1000 class labels. 
The conditional probability distribution  over object categories $c$ 
can be defined using a softmax function as given below: 
\[P({y=c}|{\bf z})=\frac{exp(z_k)}{\sum_j exp(z_j)} \] 
where, in our case, ${\bf z}$ is the output of the last fully connected layer (i.e., the FC$_8$ layer). 


\subsection{Tag Features:}

\subsubsection{Automatic Image Annotation Using Deep Features:}
Not all images on social networking sites have tags associated with them or the set of tags is very sparse  \cite{mum12}. Thus, we use an auto-annotation technique to provide tags to these images based on their visual content.
For automatic image annotation, we predict top $K$ object categories from the probability distribution extracted from the deep neural network. More precisely, given an input image {\bf x}, 
we considered class labels (or object categories) for top $K$ probabilities as tags to describe an image. Figure \ref{fig:features} shows an example of tags generated by auto-annotation mechanism for $K = 5$. We can see from Figure \ref{fig:features} that the deep tags such as ``Maillot", ``Tank suit" are representative for the image, but important tags such as ``people,'' ``women" are not included. This is because the 1000 object categories used to predict the class label do not contain these tags. The user tags for the image in Figure \ref{fig:features} are: ``birthday", ``party", ``night", ``life." 

Images on social networks also give additional information about them through their comments and description section, which are usually used as tag features to predict privacy setting. For the dataset used for the privacy detection task, we found that generally user provides n-gram tags ($n\leq4$) to describe an image, but we found  several cases that used a brief description as well (almost a sentence). This brief description can be noisy and can hamper the accuracy of a system. Hence, we filtered the instances in which tags had more than 4 tokens. We removed stop words, numbers and URLs, as they do not provide any information with respect to privacy settings.

\begin{figure*}[t]
\centering
\subfigure[Private]{\includegraphics[scale=0.7]{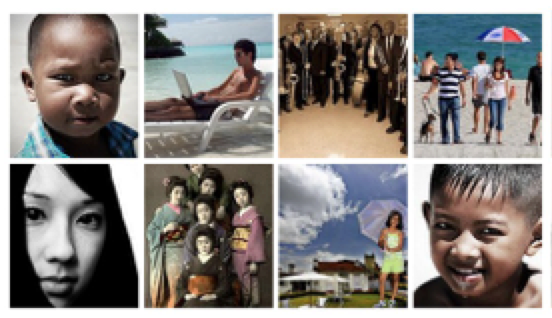}}
\subfigure[Public]{\includegraphics[scale=0.7]{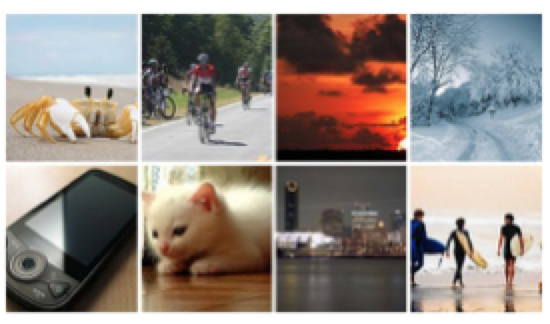}}
\caption{\label{imagesprivacy} Examples of private and public images from our PiCalert dataset.} 
\end{figure*}



\section{Dataset}
We evaluated our approach on a subset of Flickr images sampled from the PiCalert dataset (\citeauthor{Zerr:2012} \citeyear{Zerr:2012}). PiCalert consists of Flickr images on various subjects, which are manually labeled as {\em public} or {\em private} by external viewers. We randomly selected $5,000$ images from PiCalert, out of which only $4,700$ have user provided tags. These $4,700$ images were used for our privacy prediction task. The public and private images are in the ratio of 3:1. Figure \ref{imagesprivacy} shows several examples of private and public images from our dataset. 



We define an image to be {\em private} if it discloses sensitive information about a user. In Figure \ref{imagesprivacy}, images with portraits, people on the beach, family photos, etc., reveal users' personal information and, hence, are labeled as {\em private}. Public images generally depict scenery, objects, animals, etc., which do not provide any personal information about a user. However, one may wonder why an image of surfers is labeled as {\em public}. The reason is that the image is not exposing any personal information since people's faces are not visible.


\section{Experiments and Results}

In this section, we present the experimental evaluation of the proposed method. 
First, we compare deep visual features with baseline visual features, SIFT and GIST, and the SIFT and GIST  combination, for privacy prediction. 
Second, since tag features perform well on privacy prediction in previous works \cite{Squicciarini2014,Zerr:2012}, we examine the quality of tag features using both user annotated tags and auto-annotated (deep) tags. We then show an analysis of the most informative tags with respect to privacy. Finally, we evaluate the effect of using hypernym, hyponym and synonym of tags, extracted from WordNet, on the performance of privacy classifiers. 





For evaluation, we divide the PiCalert dataset into two subsets, {\bf Train} and {\bf Test}, using 6-fold stratified sampling. {\bf Train} consists of five folds randomly selected from the six folds, whereas {\bf Test} consists of the remaining fold. The number of images in {\bf Train} and {\bf Test} are $3,917$ and $783$, respectively. Stratified sampling ensures that both {\bf Train} and {\bf Test} maintain the $\approx$ 3:1 ratio between public and private images. 
 

In all experiments, we use the Support Vector Machine (SVM) classifier implemented in Weka.\footnote{ http://www.cs.waikato.ac.nz/ml/weka/}
We choose the model hyper-parameters (i.e., the C parameter and the kernel in SVM) using 5-fold cross-validation experiments on {\bf Train}. We experimented with different $C$ values, 
and two kernels, linear and RBF. Precisely, the value of $C$ and the kernel that produced the best results on {\bf Train} are used to re-train a model on {\bf Train} and to evaluate it on {\bf Test}.  

\subsection{Deep Visual Features vs. SIFT and GIST}

{\bf \em How does the performance of classifiers trained on deep visual features compare with that of classifiers trained on the traditional visual features SIFT and GIST?}

\begin{table}[t]
\centering
\begin{small}
\begin{tabular}{|l|c|c|c|c|}
\hline
{Features} & {Accuracy} & {F1-Measure} & {Precision} & {Recall}  \\
\hline
\hline
\multicolumn{5}{|c|}{{\bf Test} ($PiCalert_{783}$)}\\
\hline
\hline
FC6 & 81.10\% & 0.800	 & 0.801 & 0.811\\
FC7 & 81.23\% & 0.805 & 0.804 & 0.812\\
FC8 & \textbf{82.63\%} & \textbf{0.823} & \textbf{0.822} & \textbf{0.826}\\
Prob & 79.69\% & 0.794	 & 0.792 & 0.797\\
\hline
SIFT + GIST & 72.67\% & 0.661 & 0.672 & 0.727\\
\hline
\end{tabular}
\end{small}
\caption{Results for visual features.}
\label{table:imfeatures}
\end{table}

In this experiment, we contrast the deep visual features, FC$_6$, FC$_7$, FC$_8$, and ``Prob,'' with SIFT and GIST features and with the combination of SIFT and GIST. 

For SIFT, we constructed a vocabulary of 128 visual words for our experiments. We tried different numbers of visual words such as 500, 1000, etc., but we did not get significant improvement on the results on {\bf Train} using 5-fold cross-validation.
For a given image, GIST is computed by: (1) convolving the image with 32 Gabor filters at 4 scale and 8 orientations, which produces 32 feature maps; (2) Dividing the feature map into a $4\times4$ grid and averaging feature values of each cell; (3) concatenating these 16 averaged values for 32 feature maps, which result in a feature vector of 512 ($16 \times 32$) length.

For the deep visual features, we use an already trained deep convolutional neural network implemented in CAFFE (\citeauthor{Jia:2014:CCA:2647868.2654889} \citeyear{Jia:2014:CCA:2647868.2654889}), which is an open-source framework for deep neural networks. CAFFE implements an eight-layer network pre-trained on the ILSVRC-2012 object classification subset of the ImageNet dataset (\citeauthor{ILSVRC15} \citeyear{ILSVRC15}). The ILSVRC-2012 object classification subset consists of more than one million images annotated with 1000 object categories. 
We resize images in both {\bf Train} and {\bf Test} to the CAFFE convolutional neural net compatible size of $227 \times 227$ and encode each image using the three deep feature representations corresponding to the output of the layers FC$_6$, FC$_7$, FC$_8$, and ``Prob,'' which is the probability distribution obtained from FC$_8$ via softmax.




Table \ref{table:imfeatures} shows the performance (Precision, Recall, F1- Measure and Accuracy) of SVM using each deep feature type, FC$_6$, FC$_7$, FC$_8$, and ``Prob,'' in comparison with the performance of SVM using the combination of SIFT and GIST, on {\bf Test}. We do not show the performance of SIFT and GIST independently since they perform worse than their combination. 
The best parameter setting obtained on Train (in 5-fold cross-validation) was used on Test. For example, the value of C that gave the best results on Train using 5-fold cross-validation with FC$_8$ was $5$. Hence, $C=5$ was used on Test with the FC$_8$ feature.
As can be seen from Table \ref{table:imfeatures}, all deep visual features FC$_6$, FC$_7$, FC$_8$, and ``Prob'' outperform SIFT + GIST.  
FC$_8$ performs the best among the deep features and the performance improves as we go from FC$_6$ to FC$_8$. 
``Prob'' features which give the probability of the object labels also perform better than SIFT + GIST.




Next, we examine the quality of tag features and contrast auto-annotated (deep) tags with user annotated tags.

\subsection{Deep Tags vs. User Tags}
{\bf \em How do tag features perform on the privacy prediction task?}
We investigate the performance of SVM on user provided tags (user tags), auto-annotated tags (deep tags), and the combination of user tags and deep tags. 
For deep tags, we consider top 10 object labels as tags. We tried different $K$ values for the deep tags and achieved exceptional results with $K = 10$. Table \ref{table:tagfeatures} shows the results obtained from the experiments for tag features on the {\bf Test} set.

As can be seen from the table, deep tags perform slightly better than user tags, but the combination of the two outperforms both user tags and deep tags. 
This is because user tags consist of some general tags and deep tags consist of some specific tags. 
If we consider all general tags, then it may happen that tags overlap in two different privacy settings. For example, instead of ``swimming suit" tag if we have a more general tag as ``clothes" then this tag can appear in both privacy setting and we will not be able to differentiate between the two classes. Conversely, if we consider more specific tags then it will further differentiate between the same classes. user tags + deep tags overcomes this problem by using both the tags which gives better results.

\begin{table}[h]
\centering
\begin{small}
\begin{tabular}{|l|c|c|c|c|}
\hline
{Features} & {Acc.} & {F1-Measure} & {Prec.} & {Recall}  \\
\hline
\hline
\multicolumn{5}{|c|}{{\bf Test} ($PiCalert_{783}$)}\\
\hline
\hline
User Tags & 79.82\% & 0.782 & 0.786 & 0.798 \\
Deep Tags & 80.59\% & 0.801 & 0.799 & 0.806 \\
User + Deep Tags	& \textbf{83.14\%} & \textbf{0.827} & \textbf{0.826} & \textbf{0.831} \\
\hline
\end{tabular}
\end{small}
\caption{Results for tag features.}
\label{table:tagfeatures}
\end{table}

\begin{figure}[t]
\centering
{\includegraphics[scale=0.7]{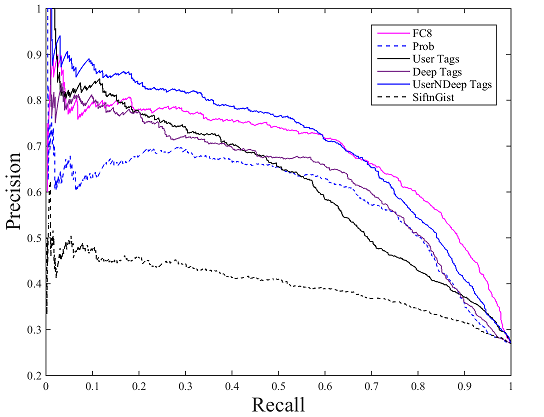}}
\caption{\label{fig:prcurve} Precision and recall curves for different features.} 
\end{figure}

Figure \ref{fig:prcurve} shows the precision-recall curves for various features for privacy prediction including baseline, FC$_8$, ``Prob,'' user tags, deep tags and user tags + deep tags. Note that the curves are shown for the private class. The precision-recall curves show that deep features and deep tags outperform the baselines. We also observe that, FC$_8$ and user tags + deep tags are the best performing approaches among all approaches described in the paper. We perform an analysis of tags to understand tags associated with privacy classes and their characteristics.

\subsection{Tag analysis}
In order to understand what tags are helpful to classify images as public or private, we explore informative tags for privacy classes and examine their characteristics. Then, we seek informative words overlapping in both private and public classes. We also study co-occurring tags for the tags that occurr in both public and private. For this analysis, we used both user provided tags and deep tags. This analysis helped us to identify tags that are correlated with privacy settings.

Informative tags help us distinguish private from public images and hence are very significant. To obtain these tags, we calculate information gain (IG) and also get the frequency of tags for the given privacy labels. In other words, tags are considered to be informative if they appear more frequently or have high information gain for a given privacy setting. 

\begin{table}
\centering
\begin{small}
\begin{tabular}{|c|c|c|}
\hline
Rank 1-5 & Rank 6-10 & Rank 11-15 \\
\hline
\textbf{Portrait}	&	Maillot			&	Bathing Cap	\\
Neck Brace			&	Wig				&	Swimming Cap	\\
Two-piece			&	Bow-tie			&	Oxygen Mask	\\
Bikini				&	\textbf{Girl}	&	Swimming Trunks \\
Tank Suit			&	\textbf{Woman}	&	Band Aid	\\

\hline
\end{tabular}
\end{small}
\caption{Tags having high information gain.}
\label{table:ig}
\end{table}

\begin{figure*}[t]
\centering
\subfigure[Private]{\includegraphics[scale=0.9]{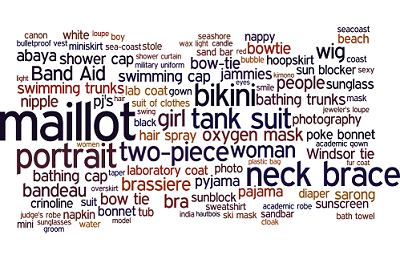}}
\subfigure[Public]{\includegraphics[scale=0.9]{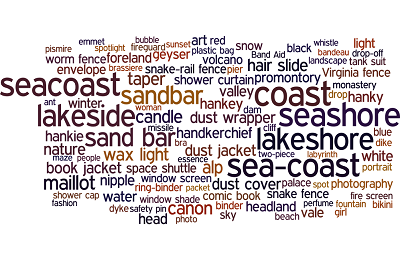}}
\caption{\label{fig:tagcloud} High frequency tag clouds with respect to public and private images.} 
\end{figure*}

High information gain shows that tags are useful to distinguish between public and private images which are in turn responsible for predicting the class labels. Table \ref{table:ig} shows tags with high information gain calculated on the {\bf Train} set, using 5-fold cross validation. Bold words indicate user provided tags, while the others are deep tags. From the table, we can conclude that deep tags contribute to a major section of IG results.
High frequency tags are the tags that are used more frequently to describe images for a given privacy setting. Thus, similar type of tags will be used for describing images with a particular privacy label. Figure \ref{fig:tagcloud} shows tag clouds for top 100 high frequency tags for private and public classes.

The tags shown with larger word size depict higher tag frequency. As we can see from the clouds, the tags depicting nature, objects such as ``sea-coast", ``lakeside", ``sandbar" etc. appear more frequently in public cloud, whereas tags describing private information such as ``portrait", ``maillot", ``bikini", ``girl" etc. appear more in private tag cloud.

\begin{figure*}[t]
\centering
\subfigure[Private]{\includegraphics[scale=0.8]{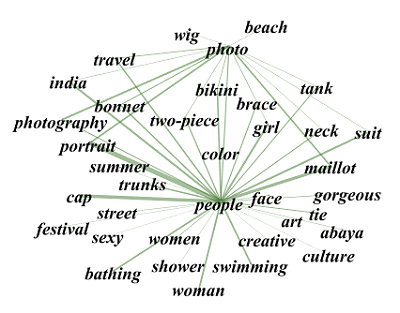}}
\subfigure[Public]{\includegraphics[scale=0.8]{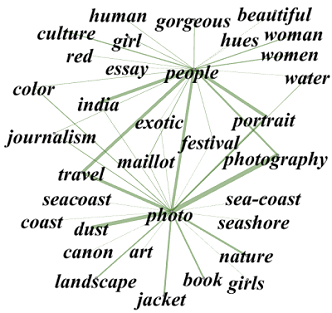}}
\caption{\label{fig:tagasso} Tag association graph.} 
\end{figure*}

We notice that there are more informative tags from the deep tags as compared to user provided tags. We also observe that there is some overlap of informative tags in public and private clouds, e.g., ``people'' and ``photo.'' Thus, we analyze other tags that co-occur with overlapping tags for further discrimination of the overlapping tags.  
It may happen that ``girl" tag can appear both in private and public images. However, ``girl" appearing with nature or natural scenes without exposing personal information will more likely represent a public category. Conversely, a ``girl" tag appearing with ``portrait" tag will denote a private category. We analyze such tags to identify co-occurring tags that assist in the privacy prediction task. To inspect these tags, we create two graphs as public and private. For the public graph, we consider each tag as a node in the graph. We draw an edge between two tags that belong to the same image. Similarly, we constructed a graph for private images.  



Figure \ref{fig:tagasso} shows a portion of the graphs for public and private images. To reduce complexity of the visualization, we display edges for only two tags viz. ``people" and ``photo". Additionally, we display only edges having co-occurrence greater than a certain threshold. Stronger edges represent high co-occurrence. We can observe from the graphs that overlapping tags such as ``people" and ``photo" tend to have different highly co-occurring tags for public and private privacy settings. Consider the ``photo" tag. ``Photo" shows high co-occurrence with ``portrait", ``maillot", ``bikini" etc. (tags describing private class) in the private graph, whereas it shows high occurrence with ``dust", ``travel", ``photograph", ``nature" etc. (tags describing public class) in the public graph. Similarly, for ``people" tag, we can see tags ``portrait", ``trunks", ``swimming" etc. in the private graph and ``photograph", ``travel", ``India" etc. in the public graph. Even if other tags in the graph are not showing very high co-occurrence, the tags occurring in the private graph tend to associate with the private class. The same is the case with the public class.


\subsection{Semantic Enrichment through WordNet}

{\bf \em Do synonyms, hyponyms, and hypernyms of the existing tags help improve the models' performance on the privacy prediction task?}
We further explore the effect of adding synonyms, hyponyms and hypernyms of the tags to existing tags. 
To analyze this, we carried out several experiments in which we examined synonyms, hyponyms and hypernyms separately, using WordNet. 
Table \ref{hyper} shows comparison of deep tags + user tags with and without hypernym.
\begin{table}
\centering
\begin{small}
\begin{tabular}{|l|c|c|c|c|}
\hline
{Features} & {Acc} & {F1-Measure} & {Prec.} & {Recall}  \\
\hline
\hline
Deep + User Tags & 83.14\% & 0.827 & 0.826 & 0.831\\
\hline
Deep + User Tags & 83.14\% & 0.822 & 0.825 & 0.831\\
Hypernym & &  & & \\
\hline
\end{tabular}
\end{small}
\caption{Comparison of results on test set using tags with and without hypernyms.}
\label{hyper}
\end{table}
We found that deep tags + user tags with and without hypernym did not have much variation on the results on {\bf Test}. 

\section{Conclusion and Future work}

In this paper, we provided a new approach for privacy prediction task. Our method was based on deep features and tag features which are proved to be exceptional than baseline approaches. We explored deep features at various network layers and also used top layer (probability) for auto-annotation mechanism. We also examined user annotated tags and deep tag features. Our experiments with integrating tag semantics to existing tags shows that existing tags provide improved results on the dataset. We also illustrated our tag related analysis with respect to privacy setting which provided us with a brief outline of the public and private images and tags associated with them. \\

We will explore the following in the future work.
\begin{enumerate}
\item Refine user provided tags by using some keyword extraction mechanism.
\item Use co-occurring words of public and private class as features for privacy classification task.
\item Combine visual features and tag features to get improved results.
\end{enumerate}


\section{Acknowledgments}
We would like to thank Adrian Silvescu and Raymond J. Mooney for helpful discussions and suggestions. 
This research is supported in part by the National Science Foundation award \#1421970  to Cornelia Caragea. Any opinions, findings, and conclusions expressed here are those of the authors and do not necessarily reflect the views of the National Science Foundation.

\nocite{}
\bibliographystyle{aaai}
\bibliography{ref}

\end{document}